# Distinct word length frequencies: distributions and symbol entropies

*Reginald Smith, Rochester, NY[1]*

**Abstract:** The distribution of frequency counts of distinct words by length in a language's vocabulary will be analyzed using two methods. The first, will look at the empirical distributions of several languages and derive a distribution that reasonably explains the number of distinct words as a function of length. We will be able to derive the frequency count, mean word length, and variance of word length based on the marginal probability of letters and spaces. The second, based on information theory, will demonstrate that the conditional entropies can also be used to estimate the frequency of distinct words of a given length in a language. In addition, it will be shown how these techniques can also be applied to estimate higher order entropies using vocabulary word length.

## 1. Introduction

The literature on word-length frequency distributions is one of the most vast amongst quantitative linguistics (cf. Best 1997, 2001; Grzybek 2006; Schmidt 1996). Word length frequencies typically investigate the frequency of words of different lengths in syllables. These distributions are common amongst texts and are typically interpreted as a type of negative binomial distribution (Altmann 1988; Wimmer & Altmann 1996) or Hyper-Poisson (Best 1998). Despite these different distributions though, they do have variations that can be used for applications such as authorship analysis (Williams 1970).

Often, the studies are conducted on the running text of a document or a corpus in order to determine the word length distribution of words within these textual sources. There is also a second tradition of word length and text analysis wherein the letter or grapheme, instead of the syllable, is used as the basic unit. This tradition has existed in parallel and has typically been used in mathematical studies in the tradition of (Shannon 1951) who analyzed the entropy of letters in texts. In this paper, we will take a limited view of word length distributions using this second tradition. In particular, we will be interested in the word length distribution only amongst distinct words in a language's vocabulary. These studies have been done in the past on the distribution of dictionary word lengths such as that in English (Rothschild 1986).

In this paper, we will investigate distinct word length distributions from two aspects. In the first part (section 2), we will investigate the typical distinct word length distribution and describe it using a derived distribution and goodness of fit tests. We will also discuss some connections between this distribution and the partition function. In the second part (sections 3-6), we will look at a different perspective based off of combinatorics to estimate the number of distinct words. Instead of the typical estimation of n-gram combinations being based only on the zero or first order entropy, using higher order conditional n-gram entropies can provide a reasonable approximation to the actual distribution.

---

[1] Address correspondence to: Reginald Smith          E-mail: rsmith@bouchet-franklin.org

## 2. Distinct word length distributions

The distinct words are calculated using files from WinEdt iSpell spell check dictionaries for a given language. Spell check dictionaries are not comprehensive by design since making them too large will include many rare or archaic words that should not be passed as correct in most writing. However, they give a good sample of commonly used terms in a language and the population of distinct words found in most texts excluding some proper nouns. The exception is the extinct language of Meroitic which used a corpus adopted from a previous paper (Smith, 2007). Below in Figure 1 is a visual comparison of the distinct word distributions by language (excluding Latin).

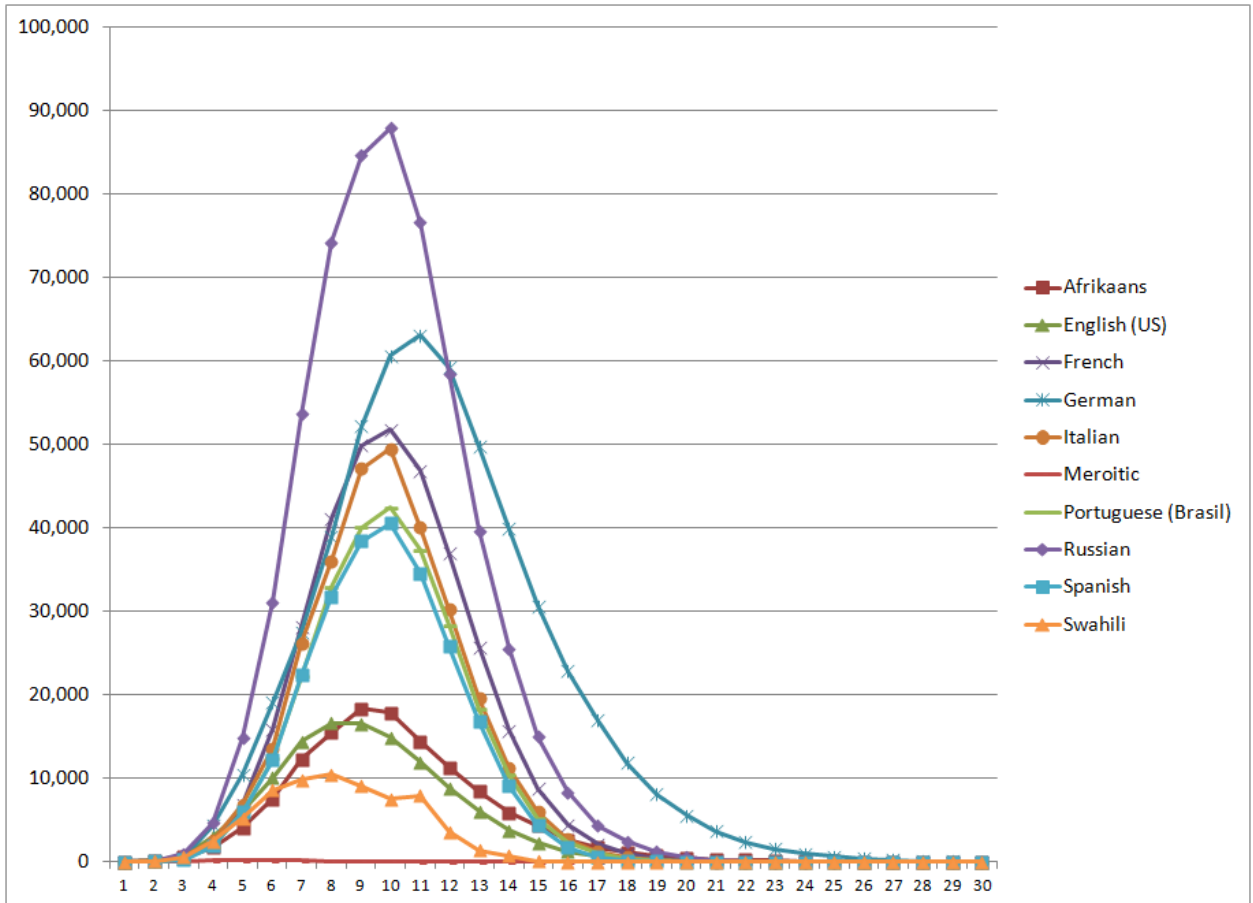

Figure 1: Graph of the frequency of distinct words by word length (Latin not shown due to a large vocabulary size that distorts the graph scale)

A key question is that given the similar nature of the distributions, is there a general distribution that can be used to describe the frequency count of distinct words in a language? If we assume the symbol set of a written language is restricted to its letters and the space/word separator character (of total symbol count $L$), a simple model of word generation can be described as follows. Assume we have a bag containing each symbol, including the space, where we can draw letters to form words and replace them after each drawing. A word is considered "finished" if you draw a space symbol from the bag. There is only one boundary condition, namely the first character must be a letter, not a space since there can be no zero-length words. Therefore, we can see that drawing from the bag can have two marginal probabilities: the probability of drawing a non-space letter (*p*) and the probability of drawing a space (" ") (*1-p*). Therefore, for a given word of length *N*, the probability of drawing a word is $p^N(1-p)$ and the probability of pulling *N* straight letters is $p^N$. Granted, this simple model ignores the fact that letters have higher order conditional probabilities for digrams, trigrams, etc. that alter the probability of letters or even spaces being subsequently drawn.

For distinct words of length *N* the virtual word length (in probabilistic terms) is $Np^N$. We term this the virtual word length since the words are obviously of length *N* but the word formation process dictates an expected value that is much lower due to the possibility of pulling a space in any one of the 1 to *N-1* selections. The virtual word length is useful since it can be used with the number of written symbols to calculate the total number of words of distinct length , $W_N$, as

$$(1) \qquad W_N = L^{Np^N} - 1$$

The final term subtracting one is to reflect the approach to zero as *N* gets large. This, using a fixed value of *p*, is similar to the geometric distribution with the last term *(1-p)* absent. However, by fitting the distribution with the correct value of *p* we can accurately model the distinct word-length distributions. In Table 1 the results are shown for calculating the *p* value that minimizes the chi-square error between the data and the distribution. The graphical results are shown in Figure 2.

In addition, there is a further observation that allows us to estimate the average word length. In particular, the average word length can be given by

$$(2) \qquad \bar{W} = \sum_{k=1}^{M} k \frac{L^{kp^k} - 1}{\sum_{j=1}^{M} L^{jp^j} - 1}$$

where *M* is the number of letters in the word of longest length. One can recognize the term in the denominator as the partition function and the average word length distribution as structurally similar to the average energy calculation in statistical mechanics. It is also the total number of distinct words, in other words, the vocabulary size.

The types of distributions represented in both equations 1 and 2 are double exponentials. Typically, closed form solutions are difficult to achieve except in the most simple of cases. Therefore, the author used numerical approximations, regressions, and simulations over various values of *p* and *L* to determine the approximate equations. From these methods, the expressions for the total vocabulary size, mean word length, and the variance of word length were found and are shown in equations 3-5:

$$(3) \qquad W_{total} = \sum_{j=1}^{M} L^{jp^j} - 1 \approx AL^{b \ln L \frac{p}{1-p}}$$

where both *A* and *b* are constants, likely the ratios of fundamental quantities.

(4) $\quad \bar{W} = -\dfrac{1}{p \ln p}$

(5) $\quad \sigma_W^2 = \dfrac{1}{[p(1-p)]^2}$

These expressions, with the constants *A* and *b* determined by minimizing chi-square error, are compared against real data in Tables 1, 2, and 3.

Table 1

The solved for *p* values for the theoretical distinct word distributions (*M*=50) along with the average word lengths (observed and based on  -1/*p* ln *p*), their percent difference, and the chi-square statistic (df=48)  and probability for the given *p* value.

| Language | Symbols | p | avg. word length (obs.) | avg. word length (exp.) | % difference | χ2 | P(χ2) |
|---|---|---|---|---|---|---|---|
| English | 27 | 0.883 | 9.2 | 9.1 | 1% | 2000 | 0 |
| Russian | 32 | 0.894 | 10.0 | 10.0 | 0% | 16176 | 0 |
| Spanish | 33 | 0.886 | 9.8 | 9.3 | 5% | 37215 | 0 |
| German | 31 | 0.893 | 11.7 | 9.9 | 18% | 284122 | 0 |
| French | 27 | 0.894 | 10.1 | 10.0 | 1% | 21204 | 0 |
| Portuguese | 27 | 0.892 | 9.9 | 9.8 | 1% | 23786 | 0 |
| Italian | 22 | 0.899 | 9.9 | 10.4 | 5% | 24482 | 0 |
| Swahili | 25 | 0.880 | 8.3 | 8.9 | 7% | 5561 | 0 |
| Afrikaans | 31 | 0.880 | 10.1 | 8.9 | 14% | 31508 | 0 |
| Latin | 24 | 0.908 | 10.9 | 11.4 | 5% | 96491 | 0 |
| Meroitic | 24 | 0.809 | 6.4 | 5.8 | 10% | 73 | 0.01 |

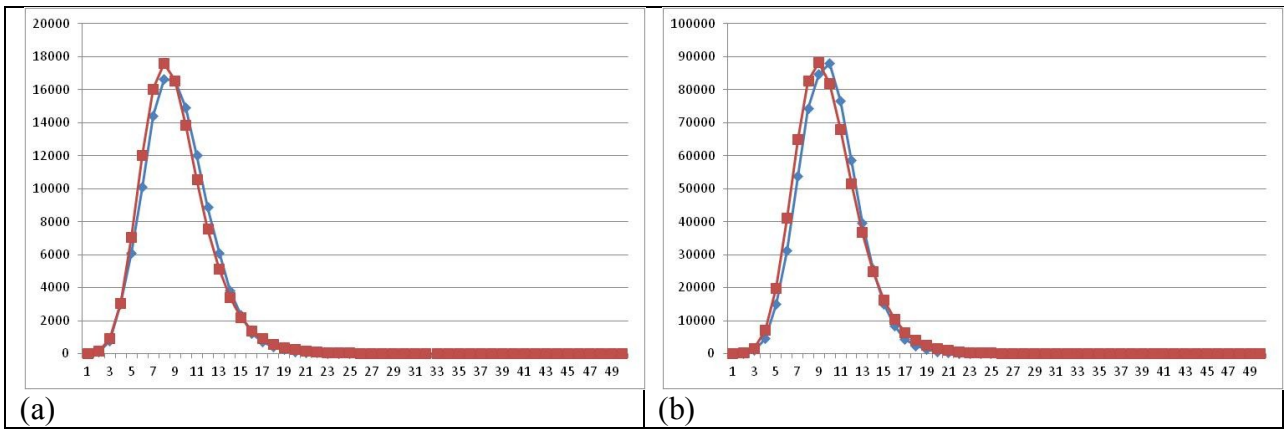

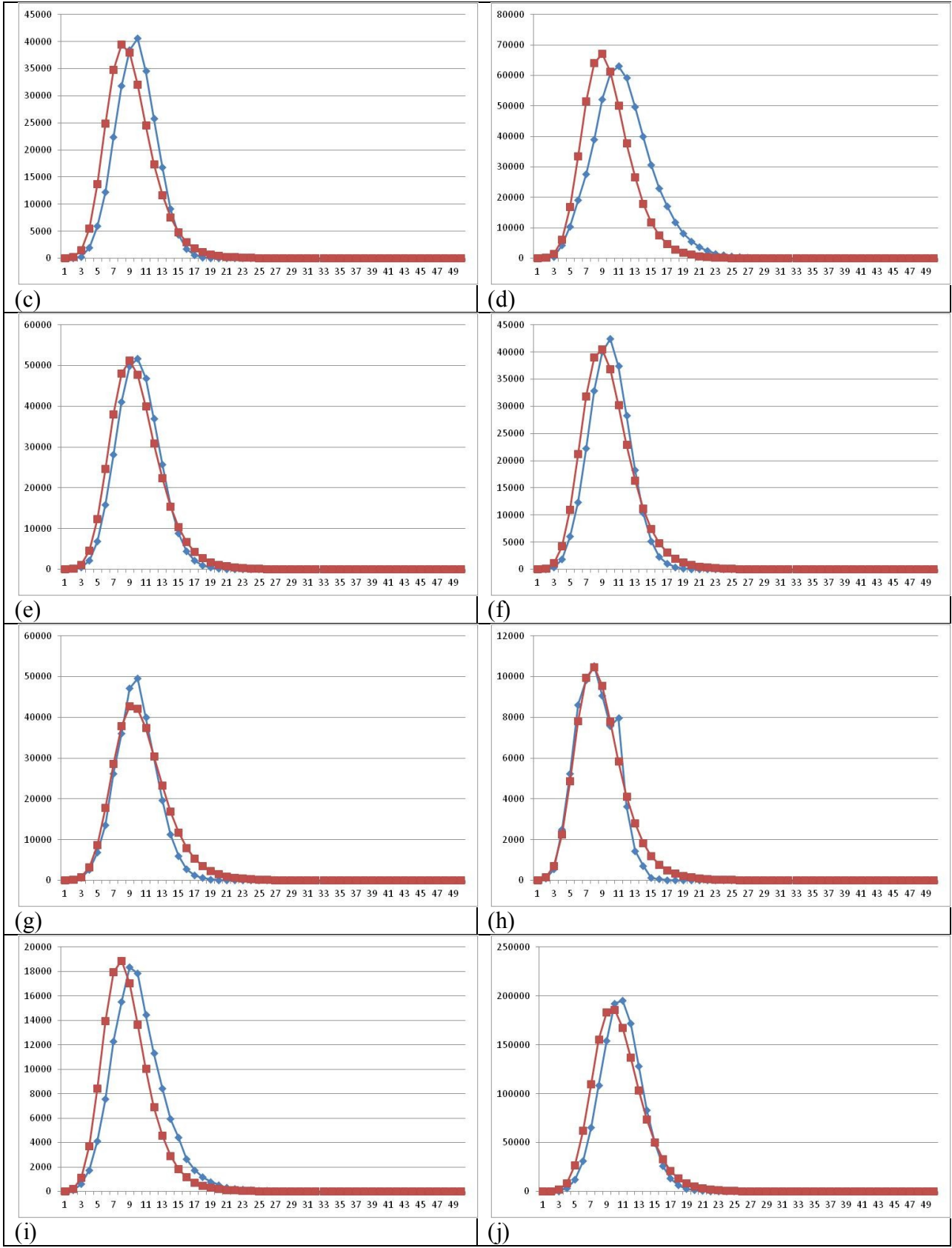

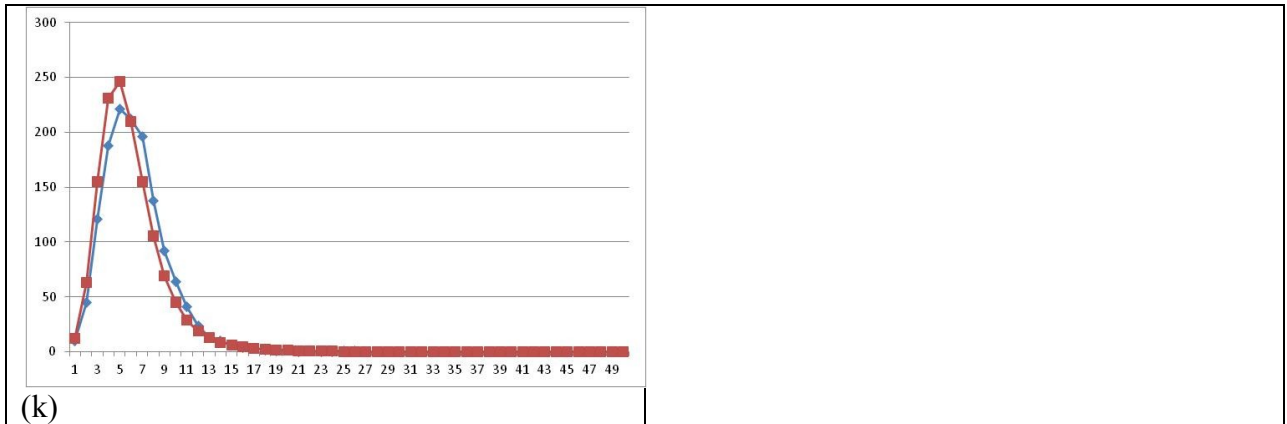

Figure 2: The distribution of distinct words by length from the spell check dictionaries (blue) and the theoretical distribution based on the *p* value in Table 1 (red). The languages shown (a-k) are English, Russian, Spanish, German, French, Portuguese, Italian, Swahili, Afrikaans, Classical Latin, and Meroitic.

Table 2
The calculated value of the standard deviation of word length using equation 5 compared to actual measured values along with the percent difference.

| Language | Symbols | p | word length $\sigma_w$ (obs.) | word length $\sigma_w$ (exp.) | % difference |
|---|---|---|---|---|---|
| **English** | 27 | 0.883 | 9.7 | 9.7 | 0% |
| **Russian** | 32 | 0.894 | 10.3 | 10.6 | 2% |
| **Spanish** | 33 | 0.886 | 10.1 | 9.9 | 2% |
| **German** | 31 | 0.893 | 12.2 | 10.4 | 17% |
| **French** | 27 | 0.894 | 10.4 | 10.6 | 1% |
| **Portuguese** | 27 | 0.892 | 10.2 | 10.4 | 2% |
| **Italian** | 22 | 0.899 | 10.2 | 11.0 | 8% |
| **Swahili** | 25 | 0.880 | 8.6 | 9.5 | 9% |
| **Afrikaans** | 31 | 0.880 | 10.6 | 9.4 | 13% |
| **Latin** | 24 | 0.908 | 11.2 | 11.9 | 7% |
| **Meroitic** | 24 | 0.809 | 7.0 | 6.5 | 9% |

Table 3
The calculated total vocabulary size along with the fitted parameters for equation 3 with a fitted
$A$=7.45.

| Language | Symbols | p | b | Vocab. size (obs.) | Vocab. size (exp.) | % difference |
|---|---|---|---|---|---|---|
| **English** | 27 | 0.883 | 0.118 | 118,619 | 118,619 | 0% |
| **Russian** | 32 | 0.894 | 0.111 | 584,929 | 584,929 | 0% |
| **Spanish** | 33 | 0.886 | 0.110 | 247,049 | 247,048 | 0% |
| **German** | 31 | 0.893 | 0.114 | 532,276 | 532,274 | 0% |
| **French** | 27 | 0.894 | 0.117 | 338,989 | 338,989 | 0% |
| **Portuguese** | 27 | 0.892 | 0.117 | 261,798 | 261,798 | 0% |
| **Italian** | 22 | 0.899 | 0.125 | 294,977 | 294,975 | 0% |
| **Swahili** | 25 | 0.880 | 0.120 | 67,988 | 67,988 | 0% |
| **Afrikaans** | 31 | 0.880 | 0.113 | 130,564 | 130,564 | 0% |
| **Latin** | 24 | 0.908 | 0.121 | 1,243,950 | 1,243,949 | 0% |
| **Meroitic** | 24 | 0.809 | 0.122 | 1,396 | 1,396 | 0% |

The distribution from equation 1 tightly fits the measured counts of distinct words for all languages. In addition, the formulas of average word length, standard deviation, and total vocabulary work reasonably well for a parameter based on such a high level measure. Finally, it is interesting to note the relatively narrow range of $p$ from 0.88 to 0.90 for all languages except Meroitic. The variable $b$ also shows a relatively narrow range of values across languages. The shorter value for Meroitic may be partially due to the corpus separating certain bound morpheme suffixes into separate words for analysis (Smith 2007) as well as a small sample size. This likely demonstrates that language content, despite the numbers of letters or sounds, is quite similar in how words are transliterated into written language.

It must be mentioned there is an inaccuracy in the calculated distinct word length distribution from equation 1 as well as the estimate in equation 3. In particular, the equation overestimates the frequency count of large words. As an example, to estimate the largest word in a language, you can use equation 1 and set $W_n = 1$. Using the parameters for English, this estimates that the longest word in English is around 43 letters. In fact, the longest non-coined word in English, according to Oxford Dictionary, is *floccinaucinihilipilification*, at 29 letters. There is a technical term *pneumonoultramicroscopicsilicovolcanoconiosis* at 45 letters, however, this is an extreme exception. Equation 1 estimates English should have 13 different words of 29 letters each. In the WinEdt spellcheck dictionary used for the empirical distinct word counts, the longest single word is 29 letters but you must drop to 24 letter words to find more than 10 distinct words. This pattern is repeated in other languages.

So equation 1 should be used with caution when calculating the frequency count of distinct words significantly larger than the mean word length. In particular, though accuracy varies by language, typically past one standard deviation greater than the mean, the frequency counts become very inaccurate and overestimate the number of distinct words. However, because words of this length are relatively rare compared to the size of the vocabulary, the overall fit for the distribution can still be strong despite these inaccuracies. For example in English words of length greater than the mean plus one standard deviation (19 letters or greater), account for about 0.5% of all distinct words.

In the next section, we will look at estimates of the distinct word count in a very different manner borrowing from the tools of information theory.

## 3. Information Theory and Linguistics

Claude Shannon, not content to sit on his laurels after his 1948 magnum opus *A Mathematical Theory of Communication*, turned his attention to the applications of information theory to human language representations, particularly written English. In (Shannon 1951), he used both statistical analysis of text and best guesses by volunteers of the next letters in fragmented texts to estimate the entropies of different orders for letters, including spaces. In (Shannon 1951) the entropy of a given order is defined as the conditional entropy of a given letter coming after an n-gram sequence.

$$(6) \qquad H_N = -\sum_{i,j} p(b_i, j) log_2 p_{b_i}(j)$$

In the above, $p(b_i, j)$ is the probability of the n-gram $b_i, j$ and $p_{bi}(j)$ is the conditional probability of the letter $j$ after the sequence $b_i$. Note, it is a common source of confusion that this is the conditional entropy for an n-gram based on n-1 symbols, not the joint entropy of n symbols.

His effort was rapidly replicated amongst many other written languages from all parts of the world. These include German (Küpfmüller 1954; Söder 1999), Russian (Lebedev & Garmaš 1959), French (Petrova et. al. 1964), Italian (Manfrino 1960; Capocelli & Ricciardi 1980), Arabic (Wanas et. al. 1976), Brazilian Portuguese (Manfrino 1970; Gomes 2007), Farsi (Darrudi, Hejazi, Oroumchian 2004) and Spanish (Guerrero & Perez 2008; Guerrero 2009). A great overview of some of these studies and the research in general is given by (Yaglom, Yaglom 1983). In addition, the symbol entropy has been used to analyze some other non-human communication repertoires such as black capped chickadees (Hailman et. al. 1985), dolphin whistles (McCowan et. al. 1999; McCowan et. al. 2002; Ferrer-i-Cancho & McCowan 2009), and humpbacked whale sounds (Doyle et. al. 2008). The additional languages of Afrikaans, Swahili, Classical Latin, and Meroitic have also been computed by the author for this paper. The Afrikaans and Swahili Corpora are due to (Scannell 2007; Roos 2009), the Latin Corpora use a selection of classical Latin texts from the Latin Library (http://www.thelatinlibrary.com) and the Meroitic corpus was constructed from transliterated texts from the *Répertoire d'épigraphie méroïtique* (REM) (Leclant, 2000). The entropy of Afrikaans was calculated using a 26 letter Latin alphabet plus a space and the accented characters ô, ê, ë, á for a total of 31 symbols. Swahili used 23 characters plus the character 'ch' and space for a total of 25 symbols. Latin used the 23 symbols of the classical Latin script plus spaces (24 characters) and Meroitic used the 24 characters of the alphabet plus the word separator character (25 characters).

In each case, the entropies, summarized in Table 5, are both a measure of the order and redundancy for a given n-gram. A lower n-gram entropy indicates increased redundancy and the n-gram entropy at any order $n >1$ must be less than or equal to the n-gram entropy of a lower order.

## 4. Word length combinatorics

In this section, we will look at the distinct word distribution from another perspective, that of the entropy of the characters. This perspective can be useful since it allows us to relate the number of distinct words to the structure of the language as defined in information theory.

Take an imaginary language with $L$ different letters. It is well known that the number of possible combinations ($W$) in a string of length $N$ is given by $W=L^N$. Granted, these are called strings instead of words because this basic equation gives no rules or bounds on which letters appear together or how many times. Real languages are much more constrained. A more accurate estimate was introduced by (Weaver & Shannon 1963) where assuming an entropy of $H(l)$ in bits, the number of possible $N$ length strings of $L$ letters is now

$$(7) \quad W = 2^{NH(l)}$$

It may seem puzzling that $L$ appears nowhere in this equation but the fact that the base is 2 and the entropy is in bits obviates its necessity. If one wants to keep $L$, they can find the same result from the below equation calculating $H(l)$ with a logarithm of base $L$:

$$(8) \quad W = L^{NH_L(l)}$$

Again, this estimate is much reduced given that this first order entropy is usually much less than the zero order entropy $H_0 = log\,L$ that returns the base equation $W=L^N$. However, can we do even better? In particular, can we fine tune our approach so that we have a relatively accurate estimate for the number of possible strings, or words, for a given word length?

The relationships between entropy, mutual information, and conditional entropy can be clearly elucidated by Venn Diagrams (Reza 1961) where

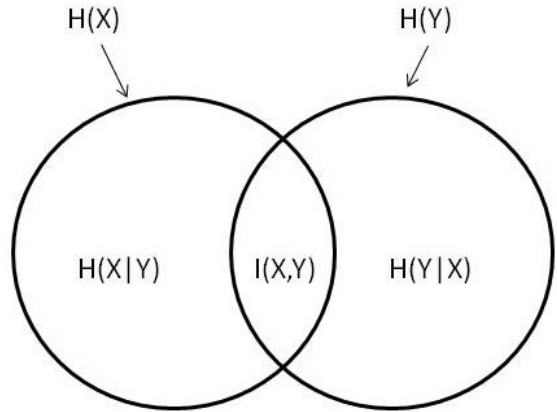

Figure 3: Venn Diagrams of entropies and mutual information

Given two symbols, the number of possible sequences of length $N$ is the total possible number given the entropy divided by those by mutual information

(9)    $W = \dfrac{2^{NH(l)}}{2^{NI}} = 2^{NH(X|Y)}$

This was an analysis first stated by Kolmogorov (Kolmogorov 1965). So the number of possible digrams given the second order conditional entropy is given by

(10)    $W_2 = 2^{2H(X|Y)}$

This can be increased for any sequence of any length, however, it becomes correspondingly inaccurate as you use equation 9 for more than two symbols. For longer sequences, the total number of possible words can be more closely estimated by the higher order conditional entropies

(11)    $W_N = 2^{NH_N}$

Of course, the "words" will depend on how the entropies are defined. If you include spaces or punctuation as symbols, this can skew the entropy and the value of $W_N$. However, for consistency with past studies, all entropies in this paper, including those in Table 5, are based only on the entropies of letters used in word formation plus the space character. Another issue, to be discussed later, is the accuracy of higher order entropies for a given text sample and how this can effect vocabulary size estimates.

Table 5
Conditional entropies up to the third order for a selection of languages.

| Language | Source | Characters | Entropy Order (Conditional Entropy) | | | |
|---|---|---|---|---|---|---|
| | | | 0 | 1 | 2 | 3 |
| English | Shannon (1951) | 27 | 4.75 | 4.14 | 3.56 | 3.30 |
| English | Schürmann and Grassberger (1996) | 27 | 4.75 | 4.08 | 3.32 | 2.73 |
| Russian | Lebedev & Garmaš (1959) | 32 | 5.00 | 4.35 | 3.52 | 3.01 |
| Spanish | Guerrero & Perez (2008); Guerrero (2009) | 33 | 5.04 | 4.15 | 3.56 | 3.09 |
| German | Söder (1999) | 31 | 4.95 | 4.06 | 3.62 | 3.25 |
| French | Petrova (1964) | 27 | 4.75 | 3.95 | 3.17 | 2.83 |
| Portuguese | Gomes (2007) | 27 | 4.75 | 3.94 | 3.56 | 3.27 |
| Portuguese | Manfrino (1970) | 23 | 4.52 | 3.91 | 3.35 | 3.20 |
| Italian | Capocelli and Ricciardi (1980) | 22 | 4.46 | 3.96 | 3.53 | 3.22 |
| Italian | Manfrino (1960) | 21 | 4.39 | 3.90 | 3.32 | 2.76 |

| Swahili | Smith - this paper (2012) | 25 | 4.64 | 3.95 | 3.33 | 2.82 |
| Afrikaans | Smith - this paper (2012) | 31 | 4.95 | 4.02 | 3.44 | 2.77 |
| Classical Latin | Smith - this paper (2012) | 24 | 4.58 | 3.90 | 3.24 | 2.79 |
| Meroitic | Smith - this paper (2012) | 24 | 4.58 | 4.24 | 3.10 | |

## 5. Results of the Entropy Method

In Table 6, a comparison of the predicted vocabulary given conditional entropy and actual numbers of distinct words of lengths two and three are given. The same corpus of Meroitic that provided the distinct words provides both the first and second order entropies. The corpus is too small to sample for third order entropy since the number of tokens is only slightly higher than 1,000 and much less than $24^3 = 13,824$.

Table 6
The predicted number of distinct tokens of length two and three determined from the conditional entropies in Table 5 and the actual number of distinct words of length two and three from WinEdt iSpell spell check dictionaries.

| Language | Source | Characters | Calculated n-grams | | Dictionary n-grams | |
|---|---|---|---|---|---|---|
| | | | 2 | 3 | 2 | 3 |
| English | Shannon (1951) | 27 | 139 | 955 | 93 | 754 |
| English | Schürmann and Grassberger (1996) | 27 | 100 | 292 | 93 | 754 |
| Russian | Lebedev & Garmaš (1959) | 32 | 132 | 523 | 87 | 995 |
| Spanish | Guerrero & Perez (2008); Guerrero (2009) | 33 | 139 | 617 | 64 | 300 |
| German | Söder (1999) | 31 | 151 | 861 | 164 | 546 |
| French | Petrova (1964) | 27 | 81 | 360 | 165 | 497 |
| Portuguese | Gomes (2007) | 27 | 139 | 898 | 76 | 338 |
| Portuguese | Manfrino (1970) | 23 | 104 | 776 | 76 | 338 |
| Italian | Capocelli and Ricciardi (1980) | 22 | 133 | 809 | 164 | 642 |
| Italian | Manfrino (1960) | 21 | 100 | 311 | 164 | 642 |
| Swahili | Smith - this paper (2012) | 25 | 101 | 352 | 95 | 557 |
| Afrikaans | Smith - this paper (2012) | 31 | 118 | 317 | 93 | 598 |
| Classical Latin | Smith - this paper (2012) | 24 | 89 | 331 | 73 | 465 |
| Meroitic | Smith - this paper (2012) | 24 | 74 | | 45 | 121 |

By comparison of Table 6, it is clear that most estimates for the number of distinct tokens in the language are if not close, of the relatively same magnitude. One of the largest issues is that the predicted value is very sensitive to the value of $NH_N$ so that variances in calculating the entropy, for example from the two Italian and Brazilian Portuguese examples, can lead to relatively large differences in the estimated numbers of distinct words.

Another possible revelation is that the higher order entropies can be equated with the expression from Section 2 as

$$(12) \quad W_N \approx 2^{NH_N} \approx L^{Np^N}$$

and by extension

$$(13) \quad H_N \approx p^N \frac{\ln L}{\ln 2}$$

Equation 13 is an estimate derived by reducing the structure of higher order entropies from $N$ classes into 2, one with probability $p$ and the other with probability $1\text{-}p$. It is obviously only an approximation and having a single value of $p$ will neglect some of the conditional probability structure of letters and spaces. However, it is an interesting connection for future work.

## 6. Reverse Estimate of Text Entropies

While the size of texts may preclude calculating higher order entropies, the techniques outlined in the previous section allow us to back out estimates of these higher order entropies based on the vocabulary size by word length. In particular, applying the reverse transformation to equation 11 we have

$$(14) \quad H_N = \frac{log_2 W_N}{N}$$

Applying this to the distinct words by length of English and German, we can estimate the higher order entropies as shown in Table 7.

Table 7

Higher order entropies estimated from the distinct words by word length.

| Order | English Words | Implied Entropy | German Words | Implied Entropy |
|---|---|---|---|---|
| 2 | 93 | 3.27 | 164 | 3.68 |
| 3 | 754 | 3.19 | 546 | 3.03 |
| 4 | 3027 | 2.89 | 4323 | 3.02 |
| 5 | 6110 | 2.52 | 10486 | 2.67 |
| 6 | 10083 | 2.22 | 19092 | 2.37 |
| 7 | 14424 | 1.97 | 27574 | 2.11 |
| 8 | 16624 | 1.75 | 38933 | 1.91 |
| 9 | 16551 | 1.56 | 52212 | 1.74 |
| 10 | 14888 | 1.39 | 60596 | 1.59 |
| 11 | 12008 | 1.23 | 63115 | 1.45 |
| 12 | 8873 | 1.09 | 59232 | 1.32 |
| 13 | 6113 | 0.97 | 49708 | 1.20 |
| 14 | 3820 | 0.85 | 39908 | 1.09 |
| 15 | 2323 | 0.75 | 30678 | 0.99 |
| 16 | 1235 | 0.64 | 22897 | 0.91 |
| 17 | 707 | 0.56 | 16978 | 0.83 |
| 18 | 413 | 0.48 | 11883 | 0.75 |
| 19 | 245 | 0.42 | 8158 | 0.68 |
| 20 | 135 | 0.35 | 5584 | 0.62 |
| 21 | 84 | 0.30 | 3684 | 0.56 |
| 22 | 50 | 0.26 | 2398 | 0.51 |
| 23 | 23 | 0.20 | 1557 | 0.46 |
| 24 | 16 | 0.17 | 974 | 0.41 |
| 25 | 9 | 0.13 | 633 | 0.37 |
| 26 | 4 | 0.08 | 386 | 0.33 |
| 27 | 2 | 0.04 | 237 | 0.29 |
| 28 | 1 | 0.00 | 128 | 0.25 |
| 29 | 0 | 0.00 | 95 | 0.23 |
| 30 | 0 | 0.00 | 44 | 0.18 |

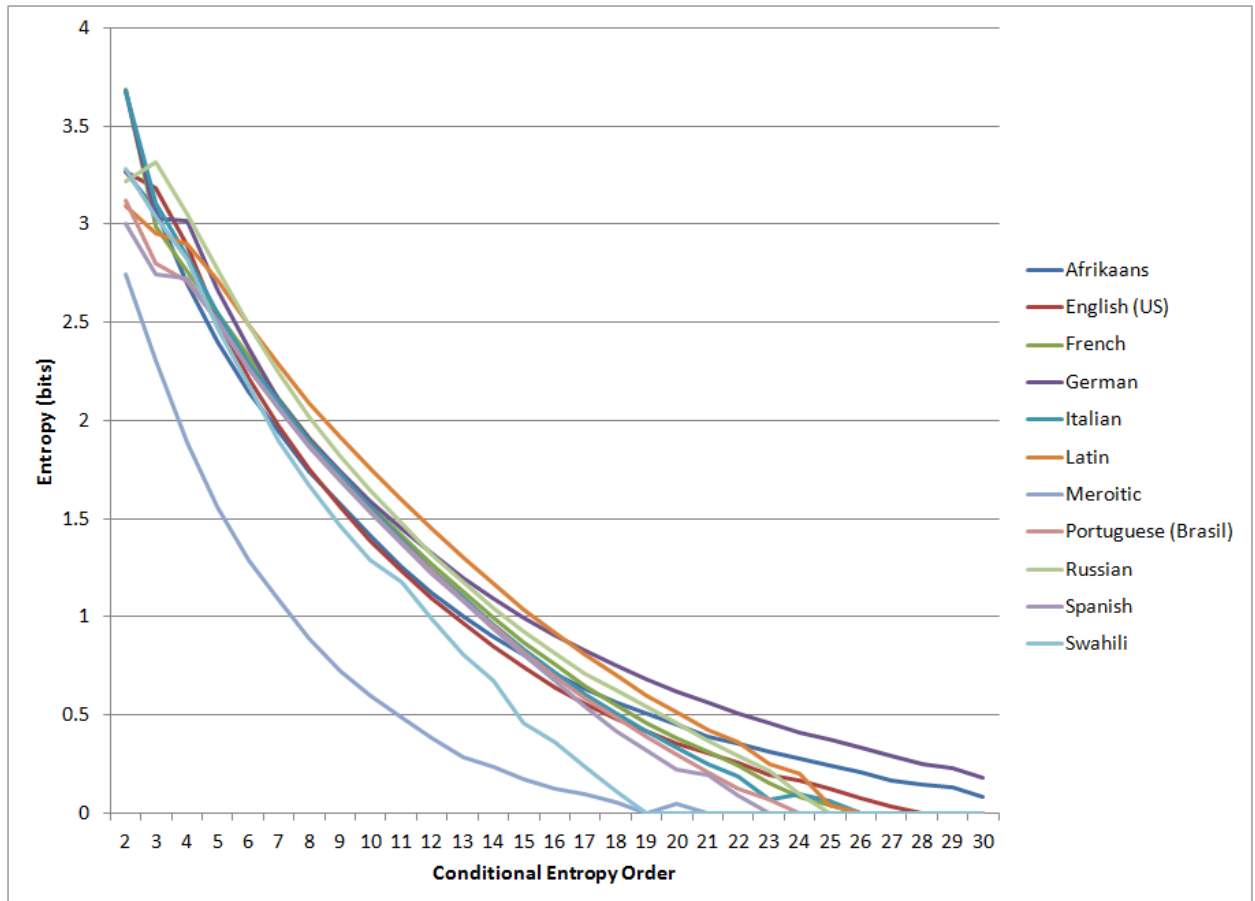

Figure 4: Graph of estimated higher order entropies by language. Meroitic is noticeably lower, likely due to the small sample size of the actual vocabulary.

**Discussion**

The results previously discussed have presented a two new lines of inquiry in the relationship between the number of distinct words of a given length and the underlying languages. First, we show that a simple distribution similar based on the number of distinct characters and the probability of a given character being a letter or a space can closely approximate the empirical distinct word length frequency distribution. Second, a relationshp between n-gram entropies and the number of distinct words of written languages was explained. Though it cannot be exact, the conditional entropy can provide a useful tool to estimate the scope of a given language by word length without a full sample of the vocabulary. Conversely, a knowledge of distinct words by length can possibly be used to estimate higher order entropies whose calculations are rendered difficult due to the massive size of the required corpus. Granted, the results can depend on which characters are used to calculate the n-gram entropy, whether spaces or punctuation are included, and other factors. One problem in the paper that is difficult to reconcile is error estimates since letter or word frequencies do not converge with a larger sample size but fluctuate as is typical in systems with large numbers of rare events (LNRE) (Baayen 2001).